# Sentence Segmentation for Classical Chinese Based on LSTM with Radical Embedding

HAN Xu[1,2], WANG Hong-su[3], ZHANG San-qian[4], FU Qun-chao[1,2], LIU S. Jun[4]

1. School of software Engineering, Beijing University of Posts and Telecommunications, Beijing 100876, China
2. Key Laboratory of Trustworthy Distributed Computing and Service, (BUPT), Ministry of Education, Beijing 100876, China
3. Insitute of Quantitative Social Science, Harvard University, Cambridge, MA, USA
4. Department of statistics, Harvard University, Cambridge, MA, USA

**Abstract**

In this paper, we develop a low-than character feature embedding called radical embedding, and apply it on LSTM (long-short term memory) model for sentence segmentation of pre-modern Chinese texts. The dataset includes over 150 classical Chinese books from 3 different dynasties and contains different literary styles. LSTM-CRF (LSTM-conditional random fields) model is a state-of-the-art method for the sequence labeling problem. Our new model adds a component of radical embedding, which leads to improved performances. Experimental results based on the aforementioned Chinese books demonstrates a better accuracy than earlier methods on sentence segmentation, especial in Tang's epitaph texts (achieving an F1-score of 81.34%).

## 1 Introduction

Many Asian languages including Chinese do not put space between characters. It is clear that word segmentation is one of the first and foremost tasks in natural language processing for these languages. Throughout the years, researchers have made significant processes in this task [1,2,3,4]. However, state-of-the-art techniques for segmentation of Chinese texts have almost exclusively focused on modern Chinese.

Pre-modern Chinese, or classical Chinese, refers to recorded Chinese texts from 1600 BC to 1800 AD. These texts contain rich information on Chinese language, literature and history. With rapid development of optical character recognition (OCR) techniques, many classical Chinese books have been digitalized to texts. For example, the largest commercial text and image depository in China, Ding-Xiu Full Text Search Platform, contains 6 billion characters. The largest noncommercial text and image depository, the Chinese Text Project, includes 5 billion characters. This gives rise to exciting opportunities in using computational techniques to retrieve and analyze these texts. For example, the Chinese Biographical Database Project [5] systematically extracts data from historical texts and converts them into a database using regular expression and tagging [6,7]. However, even as OCR techniques mature, there are still obstacles standing between standard computational methods and texts retrieved from OCR.

First, Chinese texts have no spacing between characters or words. Second, grammars of classical Chinese differ significantly from modern Chinese. Hence, commonly used corpus based on modern Chinese is not useful in processing classical Chinese texts. Third, individual characters have a richer set of meanings in classical Chinese than in modern ones, which makes it more ambiguous to define "words" from combinations of characters linguistically. As a result, the word segmentation task for classical Chinese texts is more difficult and less well defined than that for modern Chinese. Fourth, classical Chinese texts seen in historical records have no punctuations at all, whereas [8] argued that punctuation has very important meanings in Natural Language Processing (NLP), especially in Chinese. Figure 1 is an example of classical Chinese. This introduces a severe problem of sentence segmentation (in addition to word segmentation) for processing raw classical Chinese texts from OCR.

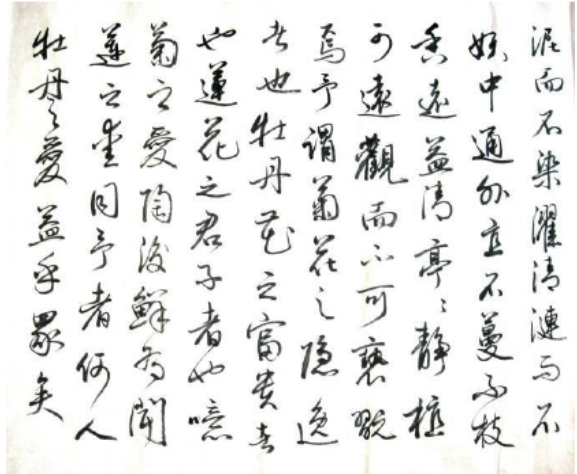

**Fig. 1** A classical Chinese texts example: it is a famous prose from Dynasty Song

In this paper, we propose a LSTM model with radical embedding to solve the pre-modern Chinese sentence segmentation problem. The main contributions of this work include that: 1) We develop an algorithm based on the Bidirectional-LSTM-CRF model with radical embedding; 2) We provide a pre-modern character embedding in a huge corpus; 3) We train a model for each dynasty separately, which results in a higher accuracy than training one model for all the text data.

## 2 Related work

Sentence boundary detection (SBD) is a language problem. The main purpose is to identify suitable breaks in sentences. Most existing research in SBD focus on the problem in the context of analyzing speech. There has not been much work in SBD for written text. The key of these problems is parts-of-speech tagging. However, for pre-modern Chinese, there's no perfect corpus with Part-of-Speech (POS) tagging for now, and it's hard to define POS in pre-modern Chinese, so it's more difficult to solve this problem.

There have been a couple of works focusing on SBD for classical Chinese in the Chinese research community. [9] proposed an algorithm to break sentences for ancient Chinese agricultural books using regular expression. They identified special syntax words and introduced punctuation around these syntax words. [10] are the first in using modern computational models for sentence segmentation in classical Chinese. Their algorithm is based on a n-gram model. [11] used a cascaded conditional random field (CRF) model, which achieved better results than [10]. [12] built upon CRF-based models by integrating phonetic information in Chinese. However, such information heavily depends on professional inputs. More recently, [13] used a RNN model for sentence segmentation. This is similar to our setup.

[14] proposed the Bi-LSTM-CRF model, which is almost state-of-the-art. It added CRF as final layer. For now, the most popular method is convolutional neural network (CNN)-Bi-LSTM-CRF [15]. However, they regard English letters as pixels, while in Chinese, each character is single and independent, it is hard to split.

[16] used Bi-LSTM-CRF to do Named Entity Recognition (NER) task. They also used character embedding to improve the performance. In English, each word is formed by many letters, as there is root, prefix and suffix in English words which has similar meanings in some occasions, so embedding by letters may get better results. However, in Chinese, every character is single and independent, so character embedding doesn't suit well. [17] proposed to use radical information to improve Chinese embedding. It's similar to the character embedding in English. But the results only slightly better than baseline models.

In this paper, we use Bi-LSTM-CRF model with radical embedding. Compared to work by [13], our paper use state-of-the-art model with CRF. Moreover, we add radical embedding as the input, the results are better than [13].

## 3 Model

### 3.1 Radical embedding

[18] proved that word embedding could highly improve performances in sequence tagging problems. In our model, we use vector representation of individual character as the input. In Chinese, most characters have a radical, which is analogous to prefix and suffix in English. It is often the case that characters that share the same radical have related meanings. As shown in Figure 2, radical '月' (moon, "mutated" from the character for meat) usually means body parts as a radical, such as '腿' (leg, unicode:u817f), '膊' (arm, unicode:u818a). Some simplified Chinese characters lost some radicals. For example, the upper radical of the character '雲' (cloud, unicode:u96f2) is '雨', which means rain; whereas the simplified one '云' only keeps the bottom part. Hence, radical embedding may capture more information in pre-modern Chinese by a corresponding segmentation method than in modern Chinese.

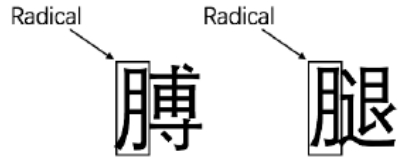

**Fig.2**　Character radical example

Every Chinese character can be represented by a unique Unicode. Characters that share the same radical are grouped together in Unicode, which can be easily retrieved in our model. The authoritative Xinhua Dictionary [19] reveals 214 radicals in total. When generating character embedding, we take these 214 radicals as parts of the input, represent them initially by randomly generated vectors, and train together as parts of the word representation. We use a Continuous Bag-Of-Words model (CBOW) for radicals, which is similar to the original CBOW model [20]. We modify the model such that each character is represented by a concatenated vector based on the character and its radical part, and maximize the log-likelihood function.

$$L = \sum_{x_i^n} \log P(x_i|h_i) \qquad (1)$$

where $x_i$ is the output Chinese character, $h_i$ is the concatenation of $c_i$ and $r_i$.

$$h_i = cat(c_{i-N}, r_{i-N}, \cdots c_{i+N}, r_{i+N}) \qquad (2)$$

where $c_i$ is the character part and $r_i$ is the radical part, N is the window size.

We make prediction using a matrix W. Different from the original CBOW model, the extra parameter introduced in the matrix W allows us to maintain the relative order of the components and treat the radical differently from the rest components.

### 3.2 Bidirectional long-short term memory

Recurrent Neural networks (RNNs) are one kind of neural networks that deal with sequential problems. In theory, RNN can handle the long-term dependencies task, but actually, it is hard to learn well when input information is too long. Also, it cannot choose important previous information. It just calculates all information without weighted. Our model follows [14]'s work, which used Bi-LSTM to solve sequence problems. LSTM includes the memory-cell in its hidden layer, which is the key of this model. The model has 3 gates: input, forget, output gate. It controls proportion of information taken from the input, the proportion of previous information to forget, and feed information to next time step. All calculations are done on the cell state. The cell receives 2 parameters $h_{t-1}$ and $c_{t-1}$ from the previous time step t-1, and input $x_t$ from the current time step. Figure 3 shows the structure of a LSTM at time step x.

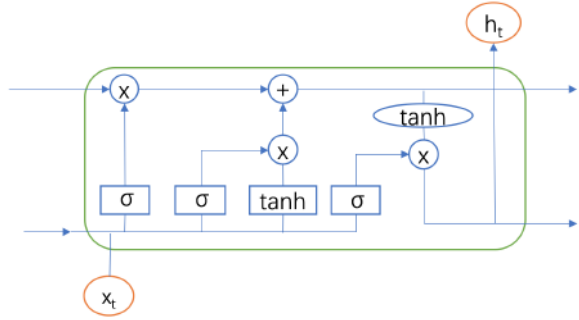

**Fig. 3**  Structure of LSTM

$$i_t = \sigma(W_{xi}x_t + W_{hi}h_{t-1} + W_{ci}c_{t-1} + b_i) \tag{3}$$

$$f_t = \sigma(W_{xf}x_t + W_{hf}h_{t-1} + W_{cf}c_{t-1} + b_f) \tag{4}$$

$$c_t = f_t c_{t-1} + i_t \tanh(W_{xc}x_t + W_{hc}h_{t-1} + b_c) \tag{5}$$

$$o_t = \sigma(W_{xo}x_t + W_{ho}h_{t-1} + W_{co}c_t + b_o) \tag{6}$$

$$h_t = o_t \tanh(c_t) \tag{7}$$

where $i_t$, $f_t$ and $c_t$ are the input, forget, and output gates, respectively. $c_t$ represents the cell state and $h_t$ denotes the hidden layer parameter at the current time step. All of these have the same dimensionality as the size of the hidden layer. σ is a standard sigmoid function, the W's are the matrices, and the b's are the bias terms.

### 3.3 CRF layer

In our model, we replace the softmax layer by a CRF layer. As shown in Figure 4, in the CRF layer, the adjacent outputs are linked each other, so we can get an optimal tagging sequence instead of an independent tagging.

$$s(x,y) = \sum_{i=0}^{n} A_{yi,yi+1} + \sum_{i=0}^{n} p_{i,yi} \tag{8}$$

The input matrix P is of size n*k, where k is the number of tagging. A is a state transition matrix, where $A_{ij}$ represents the transition probability from state i to state j. The optimal s(x,y) can be obtained by dynamic programming.

In our model, the input is a character sequence ($x_1$, $x_2$,···, $x_n$), and the output is a vector ($y_1$,$y_2$,···, $y_n$), where each $y_j$ is a probability vector corresponding to each tagging. Figure 4 shows the detail of our algorithm.

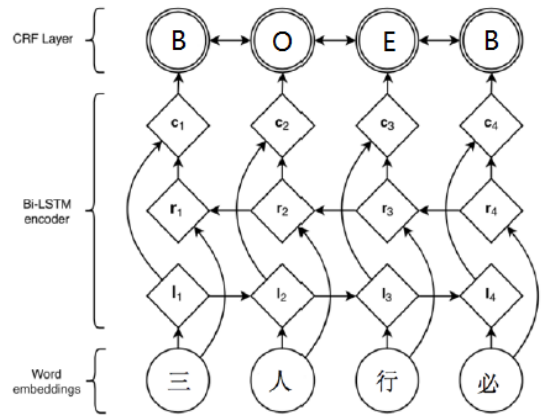

**Fig. 4** Algorithm details

## 4 Experiments

### 4.1 Data set

We obtain 150 ancient texts from [5]. The total number of characters is 44083978, and the vocabulary size is 20285. Table 1 shows the dataset details. We divide out texts according to dynasties and literary styles: Dynasty Tang (A.D.618-907), Dynasty Ming (A.D.1368-1644), Dynasty Qing (A.D.1644-1912), and Tang's epitaph. As most of the epitaphs are engraved on stones, some of the characters cannot be recognized due to corrosions caused by harsh weathers. Thus, there are a lot of unsure characters indicated by □, which is difficult to tag. To ensure sensible outputs, we deleted sentences that have more than 5 consecutive □.

Table 1. Dataset details

| Dynasty | Total Characters | Vocabulary Size |
|---|---|---|
| Tang | 6160233 | 7478 |
| Ming | 7414125 | 9952 |
| Qing | 29392770 | 10151 |
| Tang-epitaph | 1116850 | 5197 |

### 4.2 Tagging schemes

The most popular tagging method in NER tasks uses BIO format. B indicates the beginning of an entity, I means that the character is inside an entity, and O means otherwise. In this paper, we found that both of the beginning and the end of a sentence have significant feature. Inspired by BIO tags, we assign E as the end of a split sentence, B as the beginning of a split sentence, other characters are tagged by O. This tagging method is simple, and we only place punctuation marks between E and B to make sure the accuracy. Figure 5 shows a tagging result of a famous proverb of Confucius.

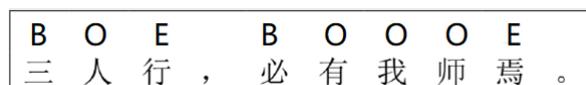

**Fig. 5** Tagging of a well-known sentence from Confucious. Translation: There is always someone, among any three people, who can teach me.

### 4.3 Preprocessing

We take the over 150 books texts as the training dataset, and obtain a classical Chinese embedding. We also train embedding by dynasties. First, we delete most punctuation marks and only keep those that mean full and half stops in the sentences. In this experiment, the punctuation set we kept is \{;:,.?!\}. Then, we separate the texts into units by dynasty. Each unit contains 100 characters, about 6-7 sentences. These units are broken up first, and split into a training set, a validation set, and a test set. The proportions are 50%, 25%, and 25%, respectively. We use Precision, Recall, and F1 to evaluate our model. We followed the [16]'s work of the hyper parameters, Table 2 shows all the parameters.

Table 2. Experiment hyper parameters

| parameter | value |
| --- | --- |
| word embedding dimension | 100 |
| hidden layer size | 100 |
| hidden layer | 1 |
| batch | 50 |
| epoch | 30 |
| learning rate | 0.01 |
| gradient clipping | 5 |
| Dropout rate | 0.5 |

$$P = \frac{TP}{TP + FP} \tag{9}$$

$$R = \frac{TP}{TP + FN} \tag{10}$$

$$F1 = \frac{2 * P * R}{P + R} \tag{11}$$

We use Precision, Recall, and F1 to evaluation our model. TP means true positive, FP means false positive, TN means true negative, FN means false negative. All the values are about the accuracy of stop signs, not tags of characters.

### 4.4 Results

We train one model for each dynasty and literary style. Table 3 shows the results on different datasets. It is seen that our new method performs the best according to the F1 score in all datasets (arranged by dynasty and literary style). All methods perform the best for Tang's epitaph. It proves that in additional to classify by dynasty, classify by literary styles can be more reasonable. Figure 6 shows the P, R and F1 value of the Tang epitaph model. Figure 7 shows the results of Tang's epitaph in different embedding method. As we can see, a unique embedding performs better than general embedding. This may be because that the language style of epitaph is different from other texts and is easier and more "standardized".

Our model performs the next best for the texts from the dynasty Qing. As the Qing has the largest dataset, and its period is the closest to modern times, the good performance of our algorithm (and CRF) may be due to both a better training and the better defined sentence structures (according to the modern Chinese rules) of the texts. Table 4 shows the example of a result of dynasty Qing, it is part of a biography about a famous poet called Dongpo hermit.

Table 3. Sentence Segmentation Results for models

| Model | Tang P | Tang R | Tang F1 | Ming P | Ming R | Ming F1 | Qing P | Qing R | Qing F1 | Tang epitaph P | Tang epitaph R | Tang epitaph F1 |
|---|---|---|---|---|---|---|---|---|---|---|---|---|
| CRF | 0.757 | 0.715 | 0.735 | 0.624 | 0.767 | 0.689 | 0.779 | 0.759 | 0.769 | 0.803 | 0.795 | 0.799 |
| LSTM | 0.766 | 0.696 | 0.729 | 0.669 | 0.672 | 0.671 | 0.686 | 0.742 | 0.713 | 0.759 | 0.781 | 0.771 |
| Bi-LSTM-CRF | 0.734 | 0.728 | 0.731 | 0.722 | 0.675 | 0.698 | 0.741 | 0.765 | 0.752 | 0.789 | 0.826 | 0.807 |
| Bi-LSTM-CRF with radical embedding | 0.747 | 0.748 | 0.748 | 0.696 | 0.714 | 0.705 | 0.761 | 0.784 | 0.772 | 0.814 | 0.813 | 0.813 |

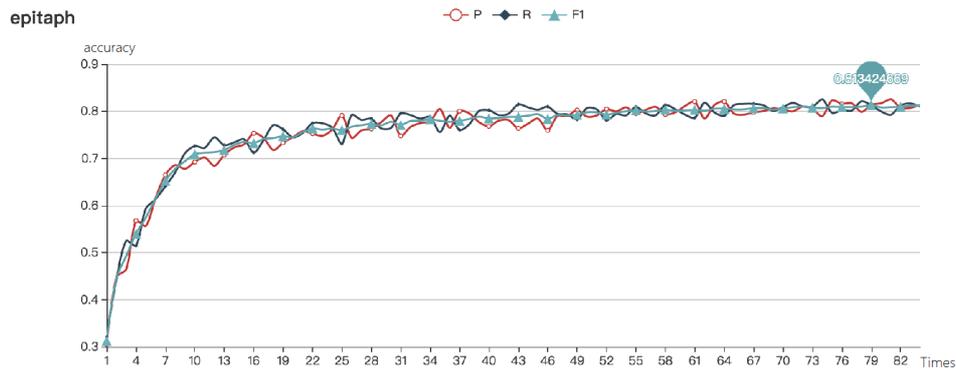

**Fig. 6** The P, R, F1 value of Tang epitaph

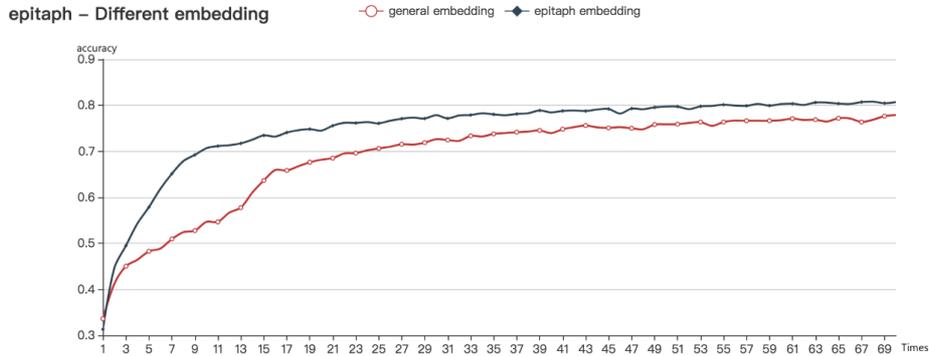

**Fig. 7** The results of general embedding and epitaph embedding on Epitaph dataset

Table 4. Case studies

| Our model: | 先生 (Prof. Su)/年四十七 (is 47 years old)/在黄州 ( in Province Huang)/寓居临皋亭 (lives in pavilion Linao)/就东坡 (near Dongpo(Location))/筑雪堂 (builds a house called Xuetang)/自号东坡居士 (called Dongpo hermit by himself)/以东坡图考之 (based on the map of Dongpo)/自黄州门南 (from the south door of Province Huang)/至雪堂四百三十步 (430 feet to Xuetang)/ |
|---|---|
| Orginal: | 先生年四十七 (Prof. Su is 47 years old)/在黄州 ( in Province Huang)/寓居临皋亭 (lives in pavilion Linao)/就东坡筑雪堂 (builds a house near Dongpo(Location) called Xuetang)/自号东坡居士 (calledDongpo hermit by himself)/以东坡图考之 (based on the map of Dongpo)/自黄州门南至雪堂四百三十步 (Xuetang is 430 feet from the south door of Province Huang)/ |

# 5 Conclusion

This paper presents a modified Bi-LSTM-CRF model with radical embedding. Our experiments show that this model outperforms existing methods in all the pre-modern Chinese text datasets we tested. The key of this new model is that we first conduct word embedding for the radicals together with the corresponding characters in the pre-training, and then use this joint embedding as the input parameter. While some earlier studies showed that including radicals do not seem to help segmenting modern Chinese texts, our study demonstrates that radicals can provide us a better handle on classical Chinese. This is consistent with the common wisdom that the "shape" of a character is more important and meaningful in pre-modern Chinese than in modern Chinese. In the future work, we hope to not only break sentences, but also tag the punctuation appropriately, which is desirable but much more challenging.

**Acknowledgments**

This work is supported by the National Key Research and Development Program of China (2017YFC1307705).